\documentclass[journal]{IEEEtran}
\usepackage[pdftex]{graphicx}
\usepackage{amsmath}

\usepackage{threeparttable}
\usepackage{booktabs}
\usepackage{color}

\usepackage[utf8]{inputenc}
\begin{document}

\title{Estimating Demand Flexibility Using Siamese LSTM Neural Networks
\thanks{G. Ruan, H. Zhong, Q. Xia, and C. Kang are with the State Key Lab of Power Systems, Department of Electrical Engineering, Tsinghua University, Beijing 100084, China.}
\thanks{D. S. Kirschen is with the Department of Electrical \& Computer Engineering, University of Washington, Seattle, WA 98195-2500 USA.}
}

\author{
Guangchun~Ruan,~\IEEEmembership{Student~Member,~IEEE,}
Daniel~S.~Kirschen,~\IEEEmembership{Fellow,~IEEE,}
Haiwang~Zhong,~\IEEEmembership{Senior~Member,~IEEE,}
Qing~Xia,~\IEEEmembership{Senior~Member,~IEEE,}
and~Chongqing~Kang,~\IEEEmembership{Fellow,~IEEE}	
}

\maketitle

\begin{abstract}
There is an opportunity in modern power systems to explore the demand flexibility by incentivizing consumers with dynamic prices. In this paper, we quantify demand flexibility using an efficient tool called time-varying elasticity, whose value may change depending on the prices and decision dynamics. This tool is particularly useful for evaluating the demand response potential and system reliability. 
Recent empirical evidences have highlighted some abnormal features when studying demand flexibility, such as delayed responses and vanishing elasticities after price spikes. Existing methods fail to capture these complicated features because they heavily rely on some predefined (often over-simplified) regression expressions. Instead, this paper proposes a model-free methodology to automatically and accurately derive the optimal estimation pattern. We further develop a two-stage estimation process with Siamese long short-term memory (LSTM) networks. Here, a LSTM network encodes the price response, while the other network estimates the time-varying elasticities. In the case study, the proposed framework and models are validated to achieve higher overall estimation accuracy and better description for various abnormal features when compared with the state-of-the-art methods.
\end{abstract}


\section{Introduction}
\subsection{Background}
\IEEEPARstart{E}{xposing} consumers to some dedicated time-varying prices, e.g., real-time prices, is regarded as a promising way to increase demand flexibility~\cite{RN22}. Massive pilot projects have proved that consumers are able to actively respond to dynamic prices~\cite{RN21}. However, these responsive behaviors may become more complicated than previously expected. For example, recent empirical evidences~\cite{RN15,RN16} discovered some nonlinear features and delayed responses after high prices. In the survey of~\cite{RN23}, participants tended to overestimate their likely demand response performance. So far, these features have not yet been clearly understood or modeled~\cite{RN6}.

Time-varying elasticity~\cite{RN52,RN53,RN54}, in this context, could serve as a fundamental tool to analyze the customers' dynamic and responsive behaviors. Compared with a constant elasticity~\cite{RN70}, the time-varying values can better capture the compounding impact of fluctuating prices and decision dynamics~\cite{RN34}, which is quite helpful to accurately assess demand flexibility and demand response potential~\cite{RN60}.

Recently, two empirically-observed phenomena, negative cross-elasticities and vanishing elasticities, presented the major challenges to deriving time-varying elasticities. According to \cite{RN28}, negative cross-elasticities might occur when a single high price reduced the demand across adjacent hours. Similar characteristic was also verified in \cite{RN15,RN16}. Real-world studies~\cite{RN26} showed that elasticities varied through the day, and consumers could be very insensitive to price in extreme conditions, leading to vanishing elasticities. Reference~\cite{RN51} also highlighted this near-zero elasticity phenomenon. Furthermore, elasticities became relatively low in~\cite{RN56} when the price was either very high or very low.

Based on these findings, a natural question that arises is how to best formulate a high-fidelity model to capture above characteristics. This paper has opted for a unique pathway by developing a model-free and data-driven methodology. Here, we specifically focus on the micro-level, short-run estimation of own- and cross-elasticities.

\subsection{Literature Review}
Elasticity estimation has long been an important topic in the literature. Reference~\cite{RN48} and \cite{RN33} provided a systematic review of the related work published on this topic, and one may easily find the following two main drawbacks.

First, most existing methods rely on some specific and (often) over-simplified regression expressions. We generally call them model-based methods, including local linear regression~\cite{RN52}, smooth time-varying coefficient cointegration approach~\cite{RN53}, dynamic logistic regression~\cite{RN11}, general McFadden form regression~\cite{RN28}, Box-Cox form regression~\cite{RN26}, two-stage least square regression~\cite{RN46,RN66}, seemingly unrelated regression~\cite{RN45}. Reference~\cite{RN49} established a hybrid equation with the use of a log-log specification, and also tested the inclusion of a squared logarithmic price term. 
Reference~\cite{RN44} tried to establish a structural model, i.e., computable general equilibrium model, but such model was often inapplicable due to many uncertain parameter settings. 
Reference~\cite{RN64} presented an inverse optimization approach to estimate the flexibility of a pool of price responsive users. The user decision model followed a given format, and the estimation could be implemented by nonlinear regression.
Different from above researches,
this paper proposes a model-free methodology to automatically and accurately derive the optimal estimation pattern from data. 

Second, existing methods often performed poorly when modeling the (complicated) temporal features. References~\cite{RN54, RN51} applied a Kalman filter approach to model the temporal features using state transition equations. In~\cite{RN15, RN16, RN36}, lagged terms are added in the regression formulas. Similarly, \cite{RN38} used a lag operator on various logarithmic terms of influential factors.
Reference~\cite{RN50} proposed a moving window to scan through the data of adjacent periods, and \cite{RN28} embedded the temporal features in a general McFadden regression expression. 
Reference~\cite{RN67} proposed a special method to apply auto-regressive expressions for error terms.
Different from above researches,
this paper chooses LSTM networks~\cite{RN1, RN3} for this estimation task because of their superior temporal modeling ability. 
In addition, we carefully develop a two-stage estimation process using two (Siamese) LSTM networks~\cite{RN55} to efficiently handle the lack of sampled elasticity data.

The success of machine learning unlocks a range of opportunities in model-free estimation~\cite{RN61}. 
Reference~\cite{RN65} used a dense neural network with Bayesian regulation backpropagation to predict the maximal demand change.
In \cite{RN68}, a dense neural network that captured the user behaviors was integrated in the model predictive control scheme.
A similar idea was applied in \cite{RN59} where some special designs were implemented to improve the estimation accuracy.
Reference~\cite{RN1,RN2} applied LSTM networks in load forecasting of residential, commercial \& industrial consumers respectively.
Then \cite{RN69} also developed the estimator with a LSTM network followed by a dynamic filter to smooth unexpected peaks with the exponential moving average.
Note that the existing works only focus on the static prediction tasks, e.g., responsive demand change~\cite{RN65}, or load forecasting~~\cite{RN1}, rather than the elasticity estimation. Following a dynamic perspective, the latter task is practically more difficult and unstable, so it is valuable to explore the potential benefit when embedding neural network techniques in elasticity estimation.

\subsection{Contributions and Paper Structure}
To the best of our knowledge, this paper provides the first model-free methodology (specifically, Siamese LSTM networks) to estimate time-varying elasticities. This method could be useful to evaluate demand flexibility, demand response potential, and system reliability. We summarize the major contributions of this paper as follows:
\begin{enumerate}
	\item The first model-free methodology to estimate time-varying elasticities is proposed in this paper, which can efficiently capture some challenging elasticity features found in real-world experiments. Our method does not rely on specific regression expressions, but can automatically and accurately derive the optimal estimation pattern from historical datasets. 
	
	\item A novel two-stage estimation process is established in this paper with Siamese LSTM networks. Within the process, a price response encoding is implemented by a LSTM network at first, and the elasticity is estimated later by another LSTM network. We connect these two networks with some tailored techniques to improve their performance in nonlinear fitting and temporal modeling.
	
	\item A rolling decision model is developed to better explain the elasticity features. This model is able to demonstrate some abnormal phenomena that have been empirically observed, including delayed responses and vanishing elasticities after price spikes.
\end{enumerate}

The rest of this paper is organized as follows. Section~\ref{SEC:FRAMEWORK} demonstrates the system framework, whose technical details are next shown in Section~\ref{SEC:ROLL} and \ref{SEC:ELSESTIMATE}. Here, the proposed two-stage elasticity estimation process with Siamese LSTM networks is fully described in Section~\ref{SEC:ELSESTIMATE}. Various case studies are analyzed in Section~\ref{SEC:CASE} to validate the method effectiveness, and finally Section~\ref{SEC:CONCLUSION} draws the conclusions.

\section{Framework} \label{SEC:FRAMEWORK}
\subsection{Problem Statement} \label{SUBSEC:PROBLEM}
This paper intends to estimate the demand flexibility of a large consumer under dynamic pricing. The available data include the historical prices, responsive load, several weather measures, and calendar observations.

Time-varying elasticity vectors are proposed to quantify this kind of flexibility, and an elasticity vector represents a segment of the well-known elasticity matrix~\cite{RN6}. In this paper, we use this vector to offer the short-term estimations of the own- and cross-elasticities for the upcoming two hours (i.e., 8 periods of 15 minutes). 

\begin{figure}[b]
	\centering
	\includegraphics[width=0.46\textwidth]{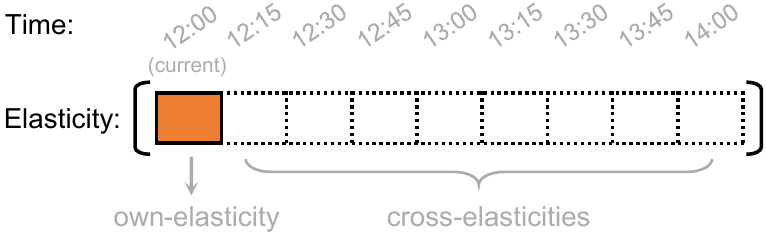}
	\caption{An illustrative elasticity vector estimated at 12 p.m. indicates demand flexibility for the next eight 15-minute periods.}
	\label{fig:vector}
\end{figure}

Fig.~\ref{fig:vector} shows an illustrative vector starting at 12 p.m. In this vector, the second element calculates the dynamic modification of power consumption at 12:15 p.m. with respect to a price change at 12 p.m.

\begin{figure*}[t]
	\centering
	\includegraphics[width=1\textwidth]{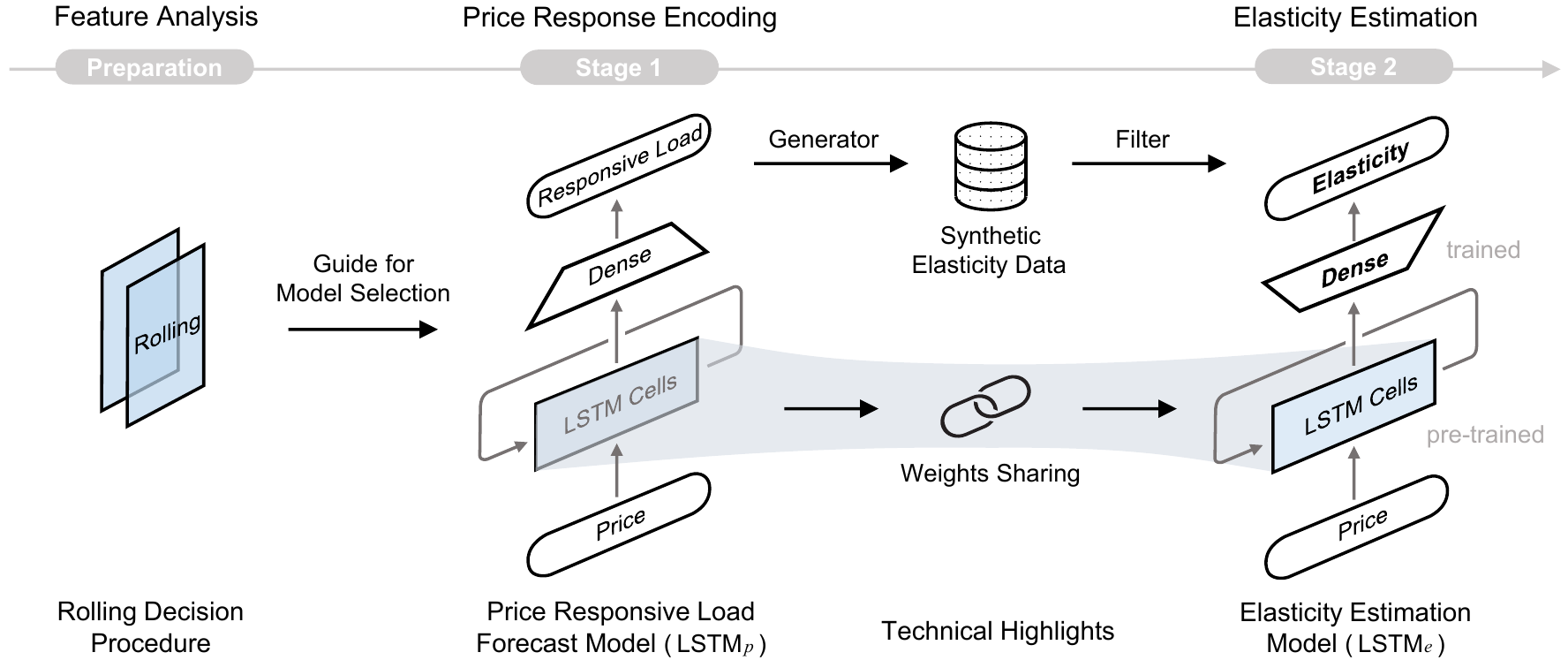}
	\caption{Proposed framework for elasticity estimation that contains the feature analysis (left) and the two-stage estimation process (middle and right).}
	\label{fig:framework}
\end{figure*}

\subsection{Overall Framework}
Fig.~\ref{fig:framework} illustrates the proposed framework that involves a preparation stage followed by a two stage estimation. 

The preparation stage involves a rolling decision procedure, which is used to elaborate the complexity of elasticity features and generate samples for the case study. Such feature analysis also lead us to choose a high-capacity model, LSTM networks, for fine-grained estimation. In particular, the two stage process involves the so-called Siamese LSTM networks, i.e., LSTM$_p$ and LSTM$_e$. In Stage~1, the responsive load data are learned by LSTM$_p$ to encode the price response features, and then in Stage~2, another network LSTM$_e$ is then established to estimate the elasticities.

Note that the two stages are specially designed to overcome the lack of sampled elasticity data when training a neural network---only responsive load data are available but our target is the elasticities.

\subsubsection{Stage 1. Price Response Encoding}
The primary goal of this stage is to create an encoding vector that fully captures the consumer's underlying features. This encoding technique follows the idea of representation learning that is generally used to boost forecasting performance~\cite{RN62}. Technically, the encoding vector is the intermediate state of LSTM$_p$, which is efficient to extract the underlying patterns in dynamic responses. More details can be found in Subsection~\ref{SUBSEC:PRE}.

\subsubsection{Stage 2. Elasticity Estimation}
As Fig.~\ref{fig:framework} shows, the Siamese LSTM networks are connected in two ways. First, inspired by photo and video interpolations~\cite{RN63}, we take advantage of LSTM$_p$ to generate some synthetic elasticity data. These data are then filtered to remove some potential outliers. Second, weights sharing allow LSTM$_e$ to access the encoding vectors created by LSTM$_p$. This means some weights of LSTM$_e$ are pre-trained in Stage~1, which can significantly reduce the training time in Stage~2. Subsection~\ref{SUBSEC:EE} discusses these points in more details.

\section{Elasticity Feature Analysis Based on the Rolling Decision} \label{SEC:ROLL}
This section analyzes the features of time-varying elasticities on a basis of rolling decisions. Our analysis reproduces and explains the empirical phenomena found in~\cite{RN15,RN16}.

\subsection{Rolling Decision Process}
\begin{figure*}[t]
	\centering
	\includegraphics[width=0.95\textwidth]{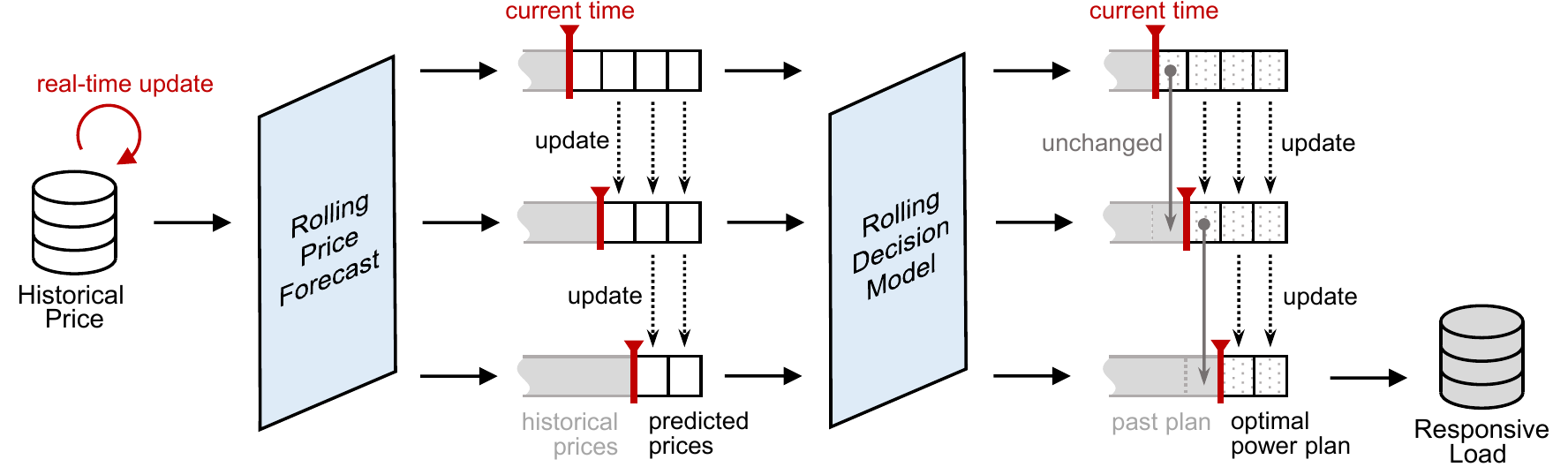}
	\caption{Rolling decision process of a large consumer in a real-time market, containing the rolling price forecast and rolling decision model.}
	\label{fig:rolling}
\end{figure*}

Consumers tend to modify their power consumption plans when the market conditions change. A rolling decision process, shown in Fig.~\ref{fig:rolling}, is widely applied in real-time markets~\cite{RN20}. Here, the consumer predicts real-time prices and repeatedly updates the prediction results. Based on these updates, the power consumption in the remaining periods is adjusted accordingly. In Fig.~\ref{fig:rolling}, the three bars after the rolling price forecast and rolling decision model have shown the decisions of three consecutive periods. Our analysis indicate that the rolling decision is influenced by both the rolling price forecast and the rolling decision model.

\subsubsection{Rolling Price Forecast}
This task is completed in two steps~\cite{RN18}: estimating the probability of price spike occurrence, and deciding to predict either the spiky prices or normal prices afterwards. Here, top 5\% prices are defined as spiky prices, and the different thresholds are allowed for different year.

We consider the mainstream methods in some highly-cited researches as a typical example. Technically, kernel SVM~\cite{RN18} is applied for predicting the probability of price spike occurrence, and two neural networks~\cite{RN19,RN17} for the prediction of spiky prices and normal prices.

\subsubsection{Rolling Decision Model}
Consider the real-time decision of a large consumer with $N$ subloads. At the current period $T_c$, this consumer will apply the following model~(\ref{eqn:rolling}) to make the power consumption plan:
\begin{align}
	\label{eqn:rolling}
	p_t = \sum_{i=1}^N F_{it} (\lambda_1, \cdots, \lambda_T, \varphi), \quad T_c \le t \le T
\end{align}
where $F_{it}$ is the decision function of subload $i$ and period $t$, which can be derived by multivariable functions or optimization models. $\lambda_t$ is the real-time price, and the prices of future time are predicted and updated by the rolling price forecast. $\varphi$ represents other input variables (optional), e.g., temperature observations. $T=96$ is the total number of period in one day. The power consumption $p_t$ is updated in a rolling manner. As time advances, the latest power consumption plan is recorded, and $T_c \leftarrow T_c+1$. The consumer will update the price prediction $\lambda_t$ again, and run model~(\ref{eqn:rolling}) again to renew the responsive load $p_t$.

Note that model~(\ref{eqn:rolling}) follows a general form and will be instantiated in specific applications. For instance, when a consumer only consider the prices of a few hours ahead, we can set the price coefficients of other periods to be zero.

\subsection{Elasticity Feature Analysis}
Elasticity describes the percentage change of power consumption in response to a percentage change of price, and the time-varying elasticity extends this concept to consider fluctuating elasticity values. Formally, the time-varying elasticity is defined as follows,
\begin{align}
	\label{eqn:els}
	e_\tau (T_c) = \frac{\text{d} p_{T_c + \tau} \ / \ p_{T_c + \tau}}{\text{d}\lambda_{T_c} \ / \ \lambda_{T_c}}
	= \frac{\text{d} p_{T_c + \tau}}{\text{d}\lambda_{T_c}} \cdot \frac{\lambda_{T_c}}{p_{T_c + \tau}}
\end{align}
where $e_\tau (T_c)$ is the elasticity that shows the dynamic demand adjustment in period~$T_c + \tau$ with respect to a price change in current period~$T_c$. 

We only consider a rolling window of $T_{rw} = 8$ periods, so an elasticity vector can be formulated as follows: 
\begin{align}
	\label{eqn:elsvector}
	E(T_c) = \left[ \ e_0 (T_c), e_1(T_c), \cdots, e_{T_{rw}}(T_c) \ \right]^\text{T}
\end{align}
where $e_0(T_c)$ is the own-elasticity, and $e_1(T_c), \cdots, e_{T_{rw}}(T_c)$ are the cross-elasticities. One could further get an own-elasticity vector $e_0$ and cross-elasticity vectors $e_1, \cdots, e_{T_{rw}}$ by concentrating the results across multiple time steps.

Combining (\ref{eqn:rolling}) and (\ref{eqn:els}), the time-varying elasticity is generally expected to be nonlinear and temporally coupled. In particular, two special features should be highlighted, i.e., vanishing elasticity effect and negative cross-elasticity effect.

\subsubsection{Vanishing Elasticity Effect}
The vanishing elasticity effect describes, by definition, the near-zero elasticity estimation for several periods after a price spike. Similar observations are demonstrated by~\cite{RN26, RN51}.

This feature can be explained with the proposed rolling decision process. Here, a consumer tends to overestimate the upcoming prices after a price spike (due to the feature of rolling price forecast, see Subsection~\ref{SUBSEC:SpecialElsFeature} for more details), and the price fluctuations on such high baselines have little impact on the final incentives. That leads to nearly no change of the consumer's decision results, indicating a tiny elasticity.

Note that this effect elaborates the short-term variation details of elasticities, different from most existing researches that simply focus on the averaged elasticities.

\subsubsection{Negative Cross-Elasticity Effect}
The negative cross-elasticity effect indicates that an increase of current price will lead to a demand drop in the future hours. This effect is consistent with the recent empirical evidences from~\cite{RN15, RN16, RN28}, but contrast to the conventional understanding~\cite{RN6}.

Conventionally, a positive cross-elasticity might be expected due to the famous compensation principle, that means when facing a price increase, a consumer will shift to use more energy in the future.

However, this is not the case when we take the consumers' decision behaviors into consideration. An important observation is that the estimated prices of different periods may greatly influence each other (not independently), especially in a rolling decision procedure. Therefore, a price increase might be followed by a higher price expectation, which results in more conservative electricity usage in the future hours. Such a bounded rationality case is similar as \cite{RN43}.

Note that this effect elaborates an abnormal variation of cross-elasticities, and could help extend the conventional knowledge.

\subsubsection{Guidance for Model Selection}
According to above discussions, estimating time-varying elasticities is a challenging task. An appropriate estimation model should have competitive advantages not only in nonlinear fitting and temporal modeling, but also in descriptive ability of special elasticity features.

However, most existing elasticity expressions are often over-simplified and may completely fail to capture the mentioned special elasticity features. This motivates us to select some model-free, high-capacity, fine-grained models for such a estimation task, e.g., learning-based models~\cite{RN61}, and the well-known LSTM networks are highly preferred~\cite{RN2}.

\section{Two-Stage Elasticity Estimation With Siamese LSTM Networks} \label{SEC:ELSESTIMATE}
This section proposes the technical details of two-stage estimation process. Here, we apply a LSTM network (LSTM$_p$) to encode price response features, and train another associated LSTM network (LSTM$_e$) to estimate elasticity vectors. We call them Siamese LSTM networks because they share the same sequential architecture and cell structure.

\subsection{Stage 1. Price Response Encoding} \label{SUBSEC:PRE}
\subsubsection{Encoding With a LSTM Model} \mbox{} \par
Price response encoding is a representation learning technique that maps price responsive features to some vectors of real numbers. These encoding vectors, compared with the raw inputs, can better extract the dominant factors, and provide a smooth manifold without wrinkles (ups and downs, due to discrete inputs).

LSTM network is an appropriate model for the above encoding task. This network is an advanced recurrent neural network with special designs to overcome the long-range dependence problem in training. Different from other ordinary neural network, a LSTM network has several cell layers to make memories, and each cell layer involves the forget, input, and output gates. Such a structure enables the temporal and dynamic behaviors, showing a full potential to capture those temporally coupled and nonlinear features of elasticities.

\begin{figure}[t]
	\centering
	\includegraphics[width=0.45\textwidth]{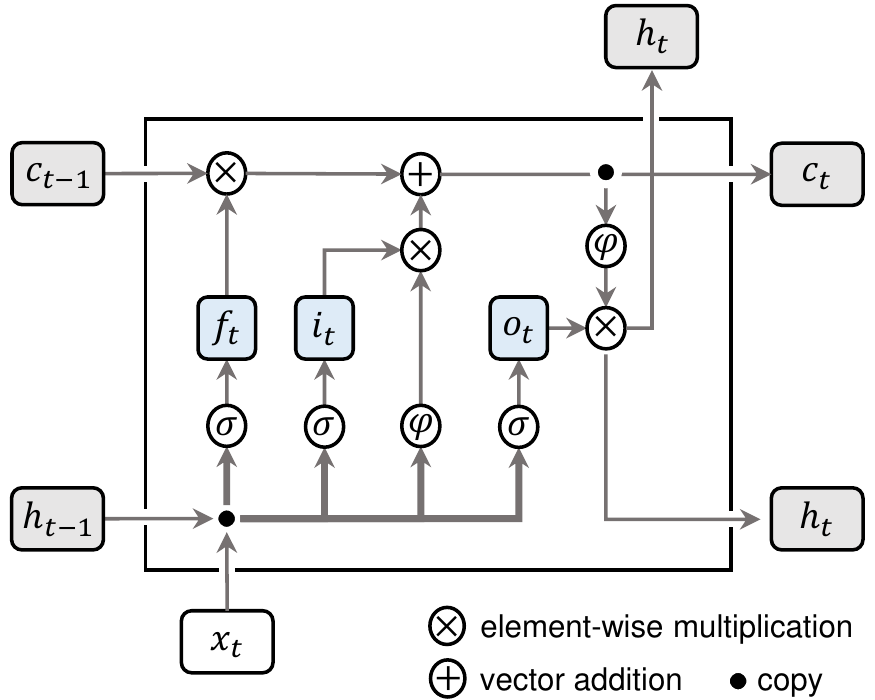}
	\caption{Internal structure of a LSTM cell layer with the forget gate $f_t$, input gate $i_t$, and output gate $o_t$.}
	\label{fig:cell}
\end{figure}

Fig.~\ref{fig:cell} shows the detailed structure of a LSTM cell layer, and it is practically verified that this structure can capture the long- and short-term features efficiently. The following formula elaborate the technical details:
\begin{subequations}
	\label{eqn:lstm}
	\begin{alignat}{5}
	f_t = \ & \sigma \left( W_f x_t + V_f h_{t-1} + b_f \right) \\
	i_t = \ & \sigma \left( W_i x_t + V_i h_{t-1} + b_i \right) \\
	o_t = \ & \sigma \left( W_o x_t + V_o h_{t-1} + b_o \right) \\
	c_t = \ & f_t \otimes c_{t-1} + i_t \otimes \varphi \left( W_c x_t + V_c h_{t-1} + b_c \right) \\
	h_t = \ & o_t \otimes \varphi (c_t)
	\end{alignat}
\end{subequations}
where $f_t$,$i_t$,$o_t$ are the outputs of forget, input, and output gates. $c_t$  is the cell state for long-term memory, $h_t$ is the intermediate state for short-term memory, and $x_t$ is the input vector. Weight matrices are denoted by $W$ and $V$ with different subscripts, and bias vectors are similarly denoted by $b$. In addition, $\sigma$ is a Sigmoid function, $\varphi$ is a Tanh function. $\otimes$ denotes the element-wise multiplication.

\begin{figure*}[t]
	\centering
	\includegraphics[width=0.98\textwidth]{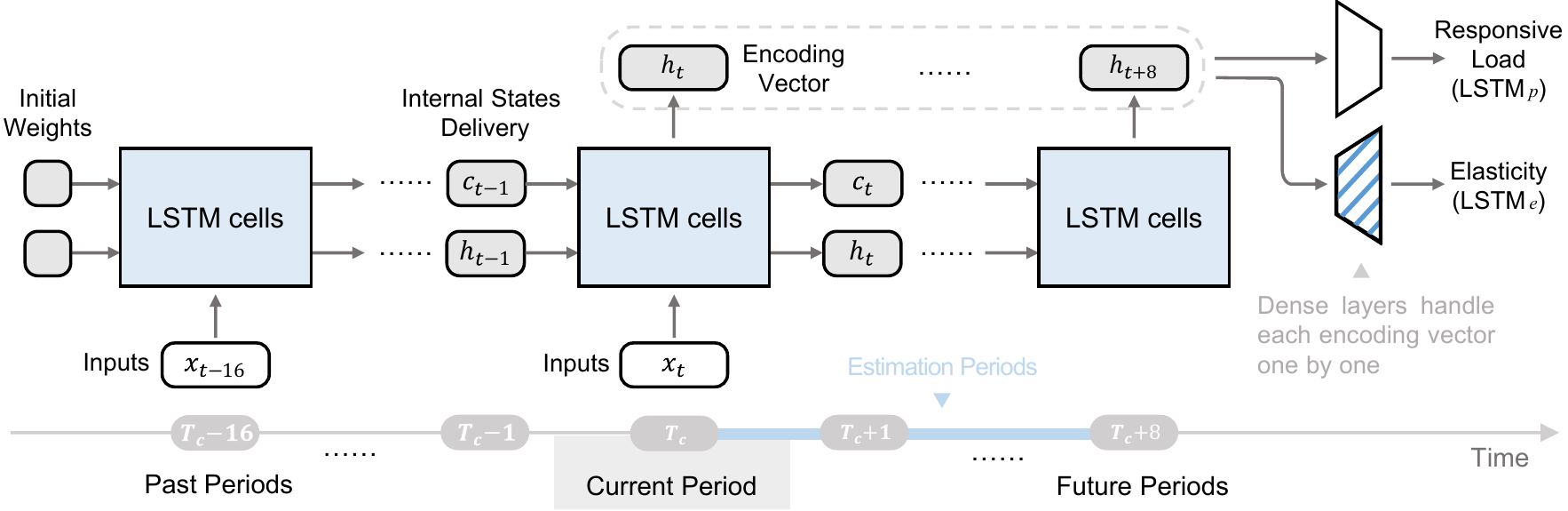}
	\caption{Unfolded sequential architecture of the proposed LSTM models (LSTM$_p$ and LSTM$_e$).}
	\label{fig:lstmpstruct}
\end{figure*}

After training, the layered and recurrent structure of a LSTM network can efficiently decode the consumer's price response features in its intermediate state $h_t$ of the last cell layer. We also call $h_t$ an encoding vector in Fig.~\ref{fig:lstmpstruct}.

\vspace{0.3em}
\subsubsection{Configurations for LSTM$_p$} \mbox{} \par
As mentioned in the Subsection \ref{SUBSEC:PROBLEM} (Problem Statement), the available data include the prices, responsive load, and other variables. LSTM$_p$ is established to understand how the responsive load is influenced by the prices and other factors. This is indeed a regression task, where the output of LSTM$_p$ is the responsive load $p_t$, and the input $x_t$ include nine specific input features, shown as follows:
\begin{itemize}
	\item Real-time price and power consumption: Continuous-valued. Prices (USD) and power consumption (MW) of past hours are both expected to be useful in capturing short-term influential factors.
	\item Weather features: Continuous-valued. Combination of the air temperature~($^{\circ}$C), relative humidity~(\%), and dew point~($^{\circ}$C). 
	\item Temporal features: Integer-valued. Combination of the period index~(1-96), weekday index~(1-7) and month index~(1-12).
	\item Holiday mark: Binary-valued. Each element can be 1 for holiday, and 0 otherwise.
\end{itemize}

Note that all input features are normalized by min-max scaling because LSTM networks are sensitive to data scales. The input $x_t$ is mathematically a concentration vector of these normalized features.

Fig.~\ref{fig:lstmpstruct} depicts the unfolded sequential architecture of the LSTM networks. We consider $T_{in}$ periods (time steps) as one sample where $T_{out}$ periods are estimated, and this paper considers $T_{out}=9$ to cover current and the upcoming 2 hours. The internal states are delivered through all periods to generate an encoding vector one after another in the last $T_{out}$ steps. There is a dense layer (white trapezoid) in Fig.~\ref{fig:lstmpstruct} to transform these encoding vectors into responsive load estimates.

LSTM$_p$ is particularly constructed with four layers (adequate for our task), involving an input layer, a single cell layer ($N_{cell}$ units), a single dense layer ($N_{den}$ units), and an output layer. For instance, ``9-28c-32-1'' means $N_{cell}=28$, $N_{den}=32$, here mark ``c'' represents the cell layer. As for the activation functions, we follow the common settings that recommend using ReLU function for the dense layer, and using Sigmoid and Tanh function for the cell layer (same as (\ref{eqn:lstm})).

The loss function of LSTM$_p$ is the popular mean squared error (MSE) function, and a mini-batch training strategy is applied so that the training process is significantly accelerated.

\subsection{Stage 2. Elasticity Estimation} \label{SUBSEC:EE}
\subsubsection{Weights Sharing} \mbox{} \par
This is a common technique to deliver some critical features between different networks. With the weights sharing, LSTM$_e$ is able to apply the encoding results learned by LSTM$_p$ in the first stage.

Applying weights sharing brings two direct benefits for LSTM$_e$: a faster training process, and a better performance. Since the shared weights in LSTM$_e$ are frozen and unchanged, the training speed can be significantly accelerated. Also, these weights consist of an efficient feature extractor that will boost the estimation performance.

\vspace{0.3em}
\subsubsection{Generator and Filter for Synthetic Elasticity Data} \mbox{} \par
A generator and a filter are proposed in order to derive some high-quality synthetic data for elasticities. We will later use these data to train the dense layer of LSTM$_e$.

The generation formula~(\ref{eqn:els-gen}) approximates (\ref{eqn:els}) by calculating the slope of a secant line near a given responsive load data with two nearby interpolated points predicted by LSTM$_p$.
\begin{align}
	\label{eqn:els-gen}
	\hat{e}_\tau (T_c) = \frac{\lambda_{T_c}}{\text{d}\lambda} \cdot
	\frac{
	\hat{p}_{T_c + \tau}^+ - \hat{p}_{T_c + \tau}^-
	}{2 \, p_{T_c + \tau}}
\end{align}
where $\hat{e}_\tau (T_c)$ is the synthetic elasticity data at the current period $T_c$, and $p_{T_c + \tau}$ is the responsive load data at period $T_c + \tau$. Further, $\hat{p}_{T_c + \tau}^+$ and $\hat{p}_{T_c + \tau}^-$ are the responsive loads interpolated by LSTM$_p$ when prices are $\lambda_{T_c} + \text{d}\lambda$ and $\lambda_{T_c} - \text{d}\lambda$ accordingly. Here, $\text{d}\lambda$ is a small price fluctuation.

The filter is used to remove the unreliable data and calculate the weighting factors for the rest. These weighting factors are important to distinguish the importance of each data point. Technically, a weighting factor $W\!F$ is derived as follows:
\begin{subequations}
	\label{eqn:weight}
	\begin{alignat}{2}
	& W\!F = \frac{1}{\eta + \alpha} \, I(\eta \ge \eta_\text{th}) \\
	& \eta = 1 - \frac{1}{T} \sum_{t=1}^T \Big( \frac{\hat{p}_t - p_t}{p_t} \Big)^2
	\end{alignat}
\end{subequations}
where $\eta$ denotes the prediction accuracy of LSTM$_p$, and $\eta_\text{th}$ is a predefined threshold. $I(\cdot)$ is an indicator function which outputs 1 if the inner condition is satisfied, and outputs 0 otherwise. Therefore, when the accuracy $\eta$ cannot reach our requirement $\eta_\text{th}$, this data will be removed as a unreliable data (get a zero weight); otherwise, the weighting factor $W\!F$ is set to be $(\eta + \alpha)^{-1}$, where $\alpha$ is an adjustment coefficient.

\begin{figure}[t]
	\centering
	\includegraphics[width=0.48\textwidth]{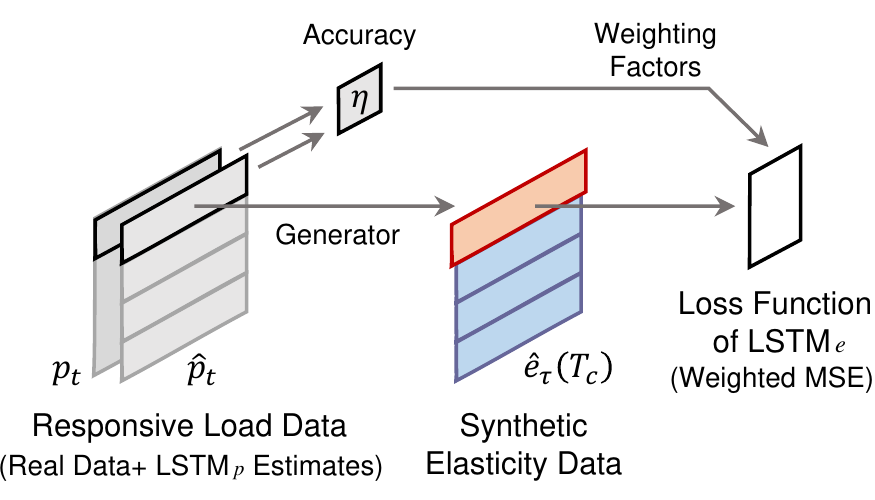}
	\caption{Illustration for the technical details in LSTM$_e$.}
	\label{fig:connection}
\end{figure}

The above technical details in LSTM$_e$ are roughly demonstrated by Fig.~\ref{fig:connection}. The functions of generator and filter is also shown in this figure.

\vspace{0.3em}
\subsubsection{Configurations for LSTM$_e$} \mbox{} \par
Fig.~\ref{fig:lstmpstruct} shows that LSTM$_e$ shares the same input data, sequential architecture, cell structure, cell weights, and activation functions as LSTM$_p$. The only differences lie in the loss function, dense layer, and output data. 

The loss function of LSTM$_e$ is configured as a weighted MSE function, with the weighting factors derived by (\ref{eqn:weight}). This is also shown in Fig.~\ref{fig:connection}.

LSTM$_e$ also applies mini-batch training, but the trainable weights are only located in the last dense layer (shaded trapezoid in Fig.~\ref{fig:lstmpstruct}). Here, the dense layer in LSTM$_e$ has no need to share the same size or weights as that in LSTM$_p$. Another difference is that the output data for training is the synthetic elasticity data.

\subsection{Overall Process for Two-Stage Estimation}
We provide a concise summary of the two-stage estimation process as follows. Given the available price and responsive load data, LSTM$_p$ is formulated in Stage~1 to capture the price response features. The weights in LSTM$_p$ is stored, and its intermediate state is extracted as an encoding vector. We use LSTM$_p$ to generate the synthetic elasticity data by (\ref{eqn:els-gen}), and the weighting factor by (\ref{eqn:weight}).

Then in Stage~2, LSTM$_e$ inherits the same cell weights of LSTM$_p$ by weights sharing technique, so the encoding vector can be assessed by LSTM$_e$ to improve the estimation performance. LSTM$_e$ further uses the weighting factors to formulate a novel loss function (weighted mean-squared-error function), and the last dense layer is trained and calibrated by the synthetic elasticity data.

After the above two-stage process, we can estimate the elasticities directly by LSTM$_e$, and don't need to run LSTM$_p$ any more. From a training perspective, the cell weights of LSTM$_e$ are (equivalently) calibrated in Stage~1, and the remaining weights are calibrated next in Stage~2. This is novel and completely different from the ordinary training patterns.

\section{Case Study} \label{SEC:CASE}
\subsection{Initial Setup}
We establish four test cases based on some real-world data, with the details shown as follows:
\begin{itemize}
	\item \textbf{Case~1}: A large commercial company in Electric Reliability Council of Texas (ERCOT) market. We collect the price data from 2006 to 2008, and use the last half year (July to December 2008) for testing. The rolling decision model is derived from~\cite{RN15}.
	
	\item \textbf{Case~2}: A building with thermal loads in Pennsylvania-New Jersey-Maryland Interconnection (PJM) market. The date range starts from April 2018 to March 2020, and the price data in 2020 are used for testing. The rolling decision model is derived from~\cite{RN60}.
	
	\item \textbf{Case~3}: An industrial company with time-shiftable loads in PJM market. The settings of this case are similar as Case~2, except for choosing another rolling decision model that is also from~\cite{RN60}.
	
	\item \textbf{Case~4}: A retail company in California market (CAISO) that serves a cluster of residential households. The price data from February 2017 to June 2019 are collected, and the last half year is considered as the testing period. Then the rolling decision model is modified from~\cite{RN36}. This case contains extreme days with persisting spiky prices or negative prices.
\end{itemize}

Note that the rolling decision models are only used to establish the data sets, and all other model details are assumed unknown when estimating the elasticity.

\begin{table}[b] 
	\centering 
	\caption{Estimation Accuracy and Breakdown Results for Different Estimation Methods} 
	\label{tab:els-acc} 
	\setlength\tabcolsep{6.5pt}  % 控制列宽 
	\begin{threeparttable} 
		\begin{tabular}{clccccc}
			\toprule
			&& SmLSTM & 2SNN  & LLR   & KFA   & GMF   \\
			\midrule
			Case~1 & RMSE      & \textbf{0.095}  & 0.216 & 0.174 & 0.108 & 0.245 \\
			& MAE       & 0.072  & 0.196 & 0.072 & \textbf{0.046} & 0.129 \\
			Case~2 & RMSE      & \textbf{0.420}  & 0.473 & 0.804 & 0.511 & 0.451 \\
			& MAE       & \textbf{0.286}  & 0.398 & 0.495 & 0.304 & 0.291 \\
			Case~3 & RMSE      & \textbf{0.195}  & 0.224 & 0.481 & 0.447 & 0.389 \\
			& MAE       & \textbf{0.133}  & 0.158 & 0.376 & 0.378 & 0.286 \\
			\midrule
			Case~1 & Own-Els   & 0.086  & 0.268 & 0.149 & \textbf{0.073} & 0.223 \\
			& Cross-Els & \textbf{0.096}  & 0.209 & 0.177 & 0.111 & 0.248 \\
			& Spike     & \textbf{0.067}  & 0.137 & 0.152 & 0.069 & 0.360 \\
			& Normal    & \textbf{0.101}  & 0.227 & 0.180 & 0.118 & 0.224 \\
			Case~2 & Own-Els   & 0.665  & \textbf{0.627} & 1.195 & 0.957 & 0.907 \\
			& Cross-Els & 0.379  & 0.450 & 0.741 & 0.424 & \textbf{0.356} \\
			& Spike     & \textbf{0.212}  & 0.311 & 0.314 & 0.283 & 0.243 \\
			& Normal    & \textbf{0.452}  & 0.496 & 0.865 & 0.548 & 0.486 \\
			Case~3 & Own-Els   & \textbf{0.277}  & 0.423 & 0.618 & 0.565 & 0.474 \\
			& Cross-Els & \textbf{0.182}  & 0.185 & 0.461 & 0.430 & 0.377 \\
			& Spike     & \textbf{0.279}  & 0.383 & 1.254 & 0.341 & 1.108 \\
			& Normal    & \textbf{0.183}  & 0.196 & 0.448 & 0.460 & 0.322 \\
			\bottomrule
		\end{tabular}
		\begin{tablenotes}
			\item Note: The overall accuracy results are evaluated by RMSE or MAE, shown in the top part; and then four breakdown results (estimation for own-elasticities, cross-elasticities, spike and normal periods) of RMSE are given in the bottom part. Most accurate items in each row are highlighted.
		\end{tablenotes}
	\end{threeparttable}
\end{table}

\begin{figure*}[t]
	\centering
	\includegraphics[width=1\textwidth]{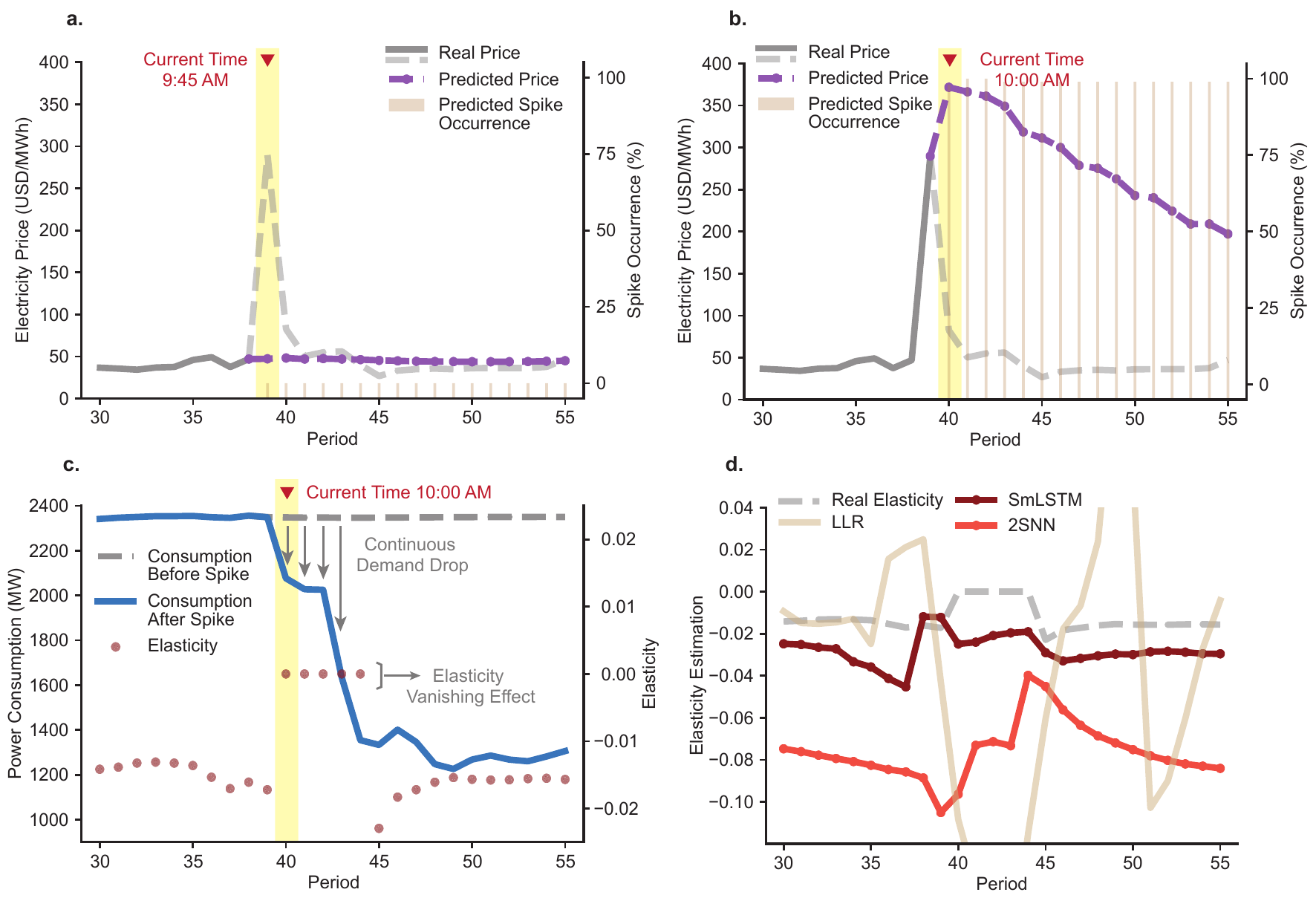}
	\caption{Illustration and estimation for the elasticity vanishing effect found in Case~1 during September 19, 2008. Subfigure (a) and (b) validate that the user may overestimate the prices after a price spike at 9:45 a.m., and then subfigure (c) shows the continuous demand drop and near-zero elasticity values. Estimation results from three typical methods are compared in subfigure (d). The zoom-in area of subfigure (d) is adjusted on purpose to cut out some abnormal points.}
	\label{fig:case-rolling}
\end{figure*}

Next, we collect four state-of-the-art methods from related work. These competing methods is helpful to demonstrate the effectiveness of the proposed model by comparison.
\begin{itemize}
	\item \textbf{SmLSTM}~(Proposed): Two-stage elasticity estimation method based on Siamese LSTM networks. The LSTM structures for price responsive load and elasticities are 9-32c-32-1, 9-32c-48-1 respectively. We use these default structures throughout the remaining discussions unless stated otherwise.
	
	\item \textbf{2SNN}: Two-stage elasticity estimation based on fully-connected neural network~\cite{RN58}. The network structures for price responsive load and elasticities are 9-32-32-9, 9-32-48-9 respectively.
	
	\item \textbf{LLR}: Local linear regression method developed in~\cite{RN52}. This formulation is a classical and popular extension of the linear regression. 
	
	\item \textbf{KFA}: Kalman filter approach from~\cite{RN54}. This method consists of a two-step calculation, and the final estimations are robust to statistical noises in general.
	
	\item \textbf{GMF}: General McFadden method from~\cite{RN28}. As a popular and representative method, it is well adapted to different estimation requirements.
\end{itemize}

Despite above configurations, we set $T_{in}=25$ (default, may change if stated), $T_{out}=9$, $\text{d}\lambda=3$~USD, $\eta_\text{th}=80\%$, $\alpha=0.5$. In addition, the mini-batch size is 256, the max iteration limit is 5000, and the day-time elasticities between period 24--80 are considered and estimated for all cases.

All simulations are running on a laptop with Intel i7-8550U CPU and 16.0 GB RAM. The programming environment is Python 3.6.0 with Tensorflow 1.12.0 and Gurobipy 8.0.0.

\begin{figure*}[t]
	\centering
	\includegraphics[width=1\textwidth]{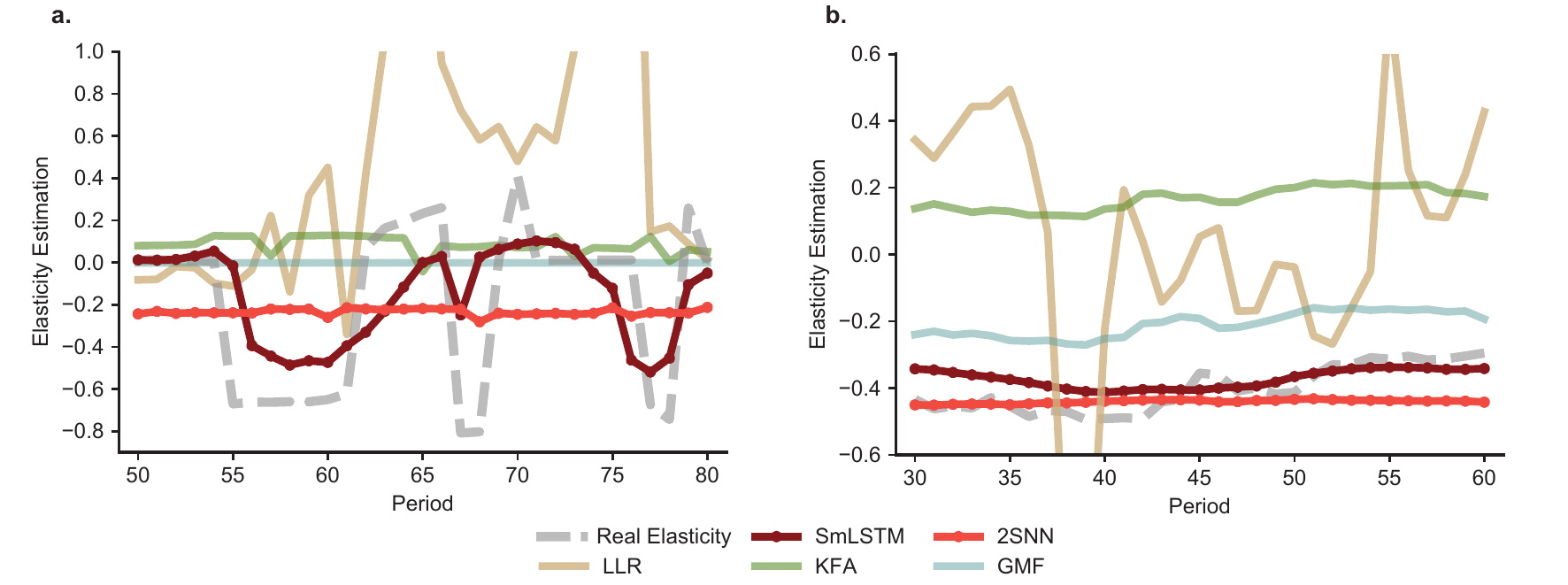}
	\caption{Estimation results for negative cross-elasticity effect. Two typical elasticities are selected for testing: $e_2$ on January 3, 2020 for Case~2 (subfigure (a)), and $e_3$ on February 1, 2020 for Case~3 (subfigure (b)). The zoom-in areas of subfigures are adjusted on purpose to cut out some abnormal points.}
	\label{fig:case-els}
\end{figure*}

\subsection{Comparison on Estimation Accuracy}
Our first focus is on the total estimation accuracy and the corresponding breakdown results for Case~1--3. Root mean square error (RMSE) and mean absolute error (MAE) are used here as the evaluation metrics.

Table~\ref{tab:els-acc} shows the estimation performance for different cases and different methods, while the total accuracy (evaluated by RMSE or MAE) and four breakdown results (estimation for own-elasticities, cross-elasticities, spike, and normal periods, all evaluated by RMSE) are given in the top and bottom parts respectively. We highlight the most accurate items in each row.

Among the three cases in Table~\ref{tab:els-acc}, SmLSTM has shown the best performance in almost all tests of total accuracy, except for the second best MAE result in Case~1. In practice, we find no large differences in using RMSE and MAE, so the breakdown results are simply calculated by RMSE. It is found that SmLSTM also performs well and balanced in different kinds of breakdown analysis. Such good and comprehensive performance makes SmLSTM superior to all other competitors.

Sharing the same two-stage process of SmLSTM, we find 2SNN estimating well in most tests except for Case~1. This is partly due to the different temporal modeling abilities, and more evidences will be given in the next subsection.

Another interesting finding in Table~\ref{tab:els-acc} is that three classical methods (LLR, KFA, and GMF) simultaneously get the most accurate results in Case~1. This is not a coincidence, and the underlying reason is that the demand model from~\cite{RN15} is simply a multivariable linear function, just roughly similar to the regression forms of LLR, KFA, and GMF. But for other model forms (e.g., Case~2 and 3), their performances are obviously deteriorated. This observation has indeed demonstrated the advantage for choosing model-free estimation methods like SmLSTM or 2SNN.

We further find that LLR, KFA, and GMF all have abnormal estimations (too large or small values), but LLR may suffer the most. More visualization results and discussions will be provided in the next subsections. SmLSTM and 2SNN, however, do not have troubles in this aspect.

\subsection{Estimation of Special Elasticity Features} \label{SUBSEC:SpecialElsFeature}
Practical applications call for detailed description of some special elasticity features. We next discuss how the competing methods are performing to detect the vanishing elasticity effect and negative cross-elasticity effect.

Fig.~\ref{fig:case-rolling} has depicted the elasticity vanishing effect. Consider the company in Case~1 who makes decisions at 9:45 a.m. and 10:00 a.m. of September 19, 2008. Comparing Fig.~\ref{fig:case-rolling}(a) and (b), a clear overestimation outcome of the following prices can be found after experiencing a sudden price spike at 9:45 a.m. These high prices incentivize the company to reduce its power consumption, as shown in Fig.~\ref{fig:case-rolling}(c). In addition, the corresponding elasticities are near-zero because the company will be very insensitive to any small price fluctuations at such a high price baseline. One can also observe the negative cross-elasticity effect in Fig.~\ref{fig:case-rolling}(c).

Note that the estimation of vanishing elasticities is much harder than those normal ones. Classical estimation methods (LLR, KFA, and GMF) will easily get abnormal estimations, and the result of LLR is provided in Fig.~\ref{fig:case-rolling}(d) as a typical example. The underlying reason is that these methods are very sensitive (even fragile) to data fluctuations in the case of tight temporal couplings, and the vanishing elasticities are unfortunately always related to such situations.

Other alternatives, SmLSTM and 2SNN, are more preferred because of their robust performance. Although not perfect, SmLSTM has provided a more reliable result in Fig.~\ref{fig:case-rolling}(d) to improve our awareness of the elasticity vanishing effect. 

Fig.~\ref{fig:case-els} has provided the estimation results for negative cross-elasticity effect. We consider two dates: January 3, 2020 for Case~2 (see Fig.~\ref{fig:case-els}(a)), and February 1, 2020 for Case~3 (see Fig.~\ref{fig:case-els}(b)). The real elasticities are plotted with grey dotted lines, and we simply compare other color lines (representing different methods) with these grey lines.

SmLSTM performs much better than other methods to accurately capture the curve fluctuation in both dates. It also shows nicer estimations in Fig.~\ref{fig:case-els}(a) than 2SNN, which is partly because of its powerful temporal modeling ability. One may also observe that LLR performs really poor in modeling negative elasticities, and KFA fails to obtain negative values.

\subsection{Effectiveness of LSTM Networks}
This subsection will dive into the implementation details of LSTM networks in Case~1--3 to better understand how they are effectively working.

\begin{figure}[t]
	\centering
	\includegraphics[width=0.48\textwidth]{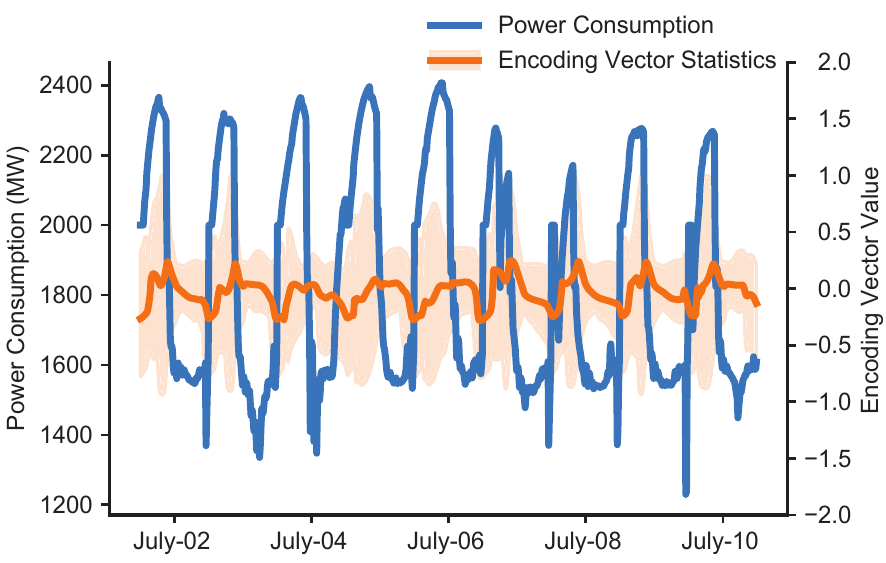}
	\caption{Visualization for encoding vectors for Case~1 during July 2--10, 2008.}
	\label{fig:case-encoding}
\end{figure}

We first concentrate on the encoding vectors learned by a LSTM network. Fig.~\ref{fig:case-encoding} visualizes the encoding vectors during July 2--10, 2008 for Case~1. Note that an encoding vector for one period has 32 elements (equal to the cell size $N_{cell}$), we thus turn to show the value range and mean value instead of the raw numbers. This establishes an orange line and a fluctuation region in Fig.~\ref{fig:case-encoding}. The mean value of encoding vectors is found to share a similar trend as the power consumption data, indicating that the LSTM cell layers have successfully captured the periodic fluctuations. This kind of temporal modeling capacity is particularly important to guarantee the good performance of SmLSTM.

Table~\ref{tab:hpopt} further discusses the robustness of the LSTM networks. Three key parameters are considered and scanned here, including the number of input steps $T_{in}$, units in the LSTM cell layer $N_{cell}$, and units in the dense layer $N_{den}$. The first row is the baseline situation, and the other simulations are established by changing one of the three key parameters. 

The direct finding from Table~\ref{tab:hpopt} is that LSTM$_e$ has a good and robust performance even when hyper-parameters are fluctuating. Although more input time steps and a larger network are beneficial to improve the estimation results, we can easily find a clear saturation effect. Therefore, a recommended strategy is starting from a large enough network, and gradually reducing the input step and network size until reaching a significant performance drop.

The last issue is the training time. Since the LSTM networks in Table~\ref{tab:hpopt} are moderate-scale, training these networks only requires less than 3.5 minutes, which is a slight burden for offline calculation.

\begin{table}[t] 
	\centering 
	\caption{Different Configurations for SmLSTM Method} 
	\label{tab:hpopt} 
	\setlength\tabcolsep{10pt}  % 控制列宽 
	\begin{threeparttable} 
		\begin{tabular}{ccccc}
			\toprule
			Input Steps & Cell Size & Dense Size & RMSE  & MAE   \\
			\midrule
			16   & 32   & 48    & 0.420  & 0.286 \\
			20   & 32   & 48    & 0.425  & 0.286 \\
			12   & 32   & 48    & 0.516  & 0.341 \\
			16   & 36   & 48    & 0.423  & 0.291 \\
			16   & 28   & 48    & 0.499  & 0.344 \\
			16   & 32   & 64    & 0.412  & 0.274 \\
			16   & 32   & 32    & 0.472  & 0.329 \\
			\bottomrule
		\end{tabular}
		\begin{tablenotes}
			\item Note: Above tests scan the number of input steps, units in the LSTM cell layer, and units in the dense layer, then evaluate the results in Case~2 by the total accuracy performance.
		\end{tablenotes}
	\end{threeparttable}	 
\end{table}

\subsection{More Extreme Tests}
We will consider Case~4 to test the model performance under the situations of persisting spiky or negative prices. In this case of CAISO market, a typical week (February 6--12, 2019) is analyzed for the impacts of persisting spiky prices, and then a typical day (May 27, 2019) for the impacts of negative prices.

Table~\ref{tab:extreme-case} summaries the accuracy performance of different estimation methods, and the statistical results are divided to demonstrate the details for the whole test periods, typical week, and typical day. As shown, the proposed method, SmLSTM, still holds a distinct advantage when comparing to other methods, especially for the days with persisting spiky prices. But its performance for negative prices seem to be average, comparable to GMF and KFA. A possible explanation is that negative prices often have a smooth (even steady) changing trend, resulting in a more predictable response of consumers~\cite{garcia2013short}---such a situation makes the advantage of integrating LSTM networks less discernible.

Nevertheless, in most electricity markets, spiky prices are much more common than negative prices, so SmLSTM is still expected to become practically useful in real-world implementations.

\begin{table}[t] 
	\centering 
	\caption{Extreme Test Results for Case~4}
	\label{tab:extreme-case} 
	\setlength\tabcolsep{5pt}  % 控制列宽 
	\begin{threeparttable} 
		\begin{tabular}{lccccc}
			\toprule
			                        & SmLSTM & 2SNN  & LLR   & KFA   & GMF   \\
			\midrule
			Whole Test              & \textbf{0.290}  & 0.539 & 0.739 & 0.496 & 0.601 \\
			Persisting Spiky Prices & \textbf{0.648}  & 0.705 & 1.273 & 1.162 & 1.238 \\
			Negative Prices         & 0.528  & 0.656 & 0.710 & 0.517 & \textbf{0.506} \\
			\bottomrule
		\end{tabular}
		\begin{tablenotes}
			\item Note: All results are estimated by RMSE. Typical days are chosen for the second and third row: February 6--12, 2019 (persisting spiky prices), and May 27, 2019 (negative prices). Most accurate items in each row are highlighted.
		\end{tablenotes}
	\end{threeparttable}	 
\end{table}

\section{Conclusion} \label{SEC:CONCLUSION}
Demand flexibility is an important but underexploited resource that contributes to balancing modern power systems. 
Major difficulties still remain in understanding and capturing consumers' adaptive responses to price signals.

The key finding of this paper is a novel modeling tool to capture consumers' dynamic behaviors when exposed to fluctuating prices. This tool is fully model-free, and involves a two-stage estimation process and Siamese LSTM networks. With these efforts, we show the first attempt to understand and estimate some real-world elasticity features.

The proposed framework and models can be readily applied for different stakeholders, e.g., retailers and distribution system operators, and also for different application areas, e.g., estimating system reliability and demand response potential.

Our future work involves in-depth investigation on the diverse features of different kind of consumers, which could provide further insights and guidance to tailor the proposed estimation models.

\bibliographystyle{ieeetr}
\bibliography{ref}

\end{document}